\documentclass[sigconf,natbib=true]{acmart}
\usepackage[T1]{fontenc}
\usepackage{latexsym}
\usepackage{booktabs}
\usepackage{multirow}
\usepackage{graphicx}
\usepackage[table]{xcolor}
\usepackage{colortbl} % for \rowcolor in tables
\usepackage{url}
\usepackage{array}
\usepackage{enumitem}
\usepackage{hyperref}
\usepackage{algorithm}
\usepackage{algpseudocode}
\usepackage{tabularx}
\usepackage{placeins} % provides \FloatBarrier to control float drift

\AtBeginDocument{%
  }

\setcopyright{acmlicensed}
\copyrightyear{2018}
\acmYear{2018}
\acmDOI{XXXXXXX.XXXXXXX}

\acmJournal{JDS}
\acmVolume{37}
\acmNumber{4}
\acmArticle{111}
\acmMonth{8}

\begin{document}

\title{EPM-RL: Reinforcement Learning for On-Premise Product Mapping in E-Commerce}

\author{Minhyeong Yu}
\orcid{0000-0002-0350-4349}
\email{minhyeong@enhans.ai}
\affiliation{%
  \institution{AI Research, Enhans}
 \state{Seoul}
  \country{South Korea}}

\author{Wonduk Seo*}\thanks{*denotes corresponding author.}
\orcid{0009-0008-6070-1833}
\email{wonduk@enhans.ai}
\affiliation{%
  \institution{AI Research, Enhans}
 \state{Seoul}
  \country{South Korea}}

\renewcommand{\shortauthors}{Trovato et al.}

\acmArticleType{Review}

\acmCodeLink{https://github.com/borisveytsman/acmart}
\acmDataLink{htps://zenodo.org/link}
%%
%% Authors' contribution
\acmContributions{BT and GKMT designed the study; LT, VB, and AP
  conducted the experiments, BR, HC, CP and JS analyzed the results,
  JPK developed analytical predictions, all authors participated in
  writing the manuscript.}

\begin{abstract}
Product mapping---the task of deciding whether two e-commerce listings refer to the same product---is a core problem for price monitoring and channel visibility. In real marketplaces, however, sellers frequently inject promotional keywords, platform-specific tags, and bundle descriptions into titles, causing the same product to appear under many different names. Recent LLM-based and multi-agent frameworks improve robustness and interpretability on such hard cases, but they often rely on expensive external APIs, repeated retrieval, and complex inference-time orchestration, making large-scale deployment costly and difficult in privacy-sensitive enterprise settings. To address these issues, we present \textbf{EPM-RL}, a reinforcement-learning-based framework for building an accurate and efficient on-premise \textbf{e-commerce product mapping model}. Our central idea is to \emph{distill high-cost agentic reasoning into a trainable in-house model}. Starting from a curated set of product pairs with LLM-generated rationales and human verification, we first perform parameter-efficient fine-tuning (PEFT) on a small student model using structured reasoning outputs. We then further optimize the model with Reinforcement Learning (RL) using an agent-based reward that jointly evaluates output-format compliance, label correctness, reasoning--preference scores from specially designed judge models. Preliminary results show that \textbf{EPM-RL} consistently improves over PEFT-only training and offers a stronger quality--cost trade-off than commercial API-based baselines, while enabling private deployment and lower operational cost. These findings suggest that reinforcement learning can turn product mapping from a high-latency agentic pipeline into a scalable, inspectable, and production-ready in-house system.
\end{abstract}


\maketitle

\section{Introduction}

Product mapping is the task of deciding whether two e-commerce listings refer to the same underlying product~\cite{aanen2015automated,yager2001product,kejriwal2021evaluation,seo2025question}. It underpins many downstream applications, including price monitoring, seller and channel visibility, catalog integration, and duplicate detection~\cite{wang2009applications,greco2018commerce}. In real marketplaces, however, the task is far from trivial: sellers often append promotional keywords, platform-specific tags, and bundle descriptions to titles, causing the same product to appear under many different names across platforms. Conversely, listings with highly similar titles can correspond to different products due to subtle differences in quantity, option type, or variant information. These challenges make robust product mapping both practically important and technically difficult~\cite{kopcke2010evaluation,primpeli2019wdc,seo2025question}.

Earlier approaches typically treated product mapping as entity matching. Traditional pipelines relied on string normalization, rule-based heuristics, token overlap, hand-crafted features, or pairwise classifiers trained on title pairs. Neural encoders later improved semantic matching by learning similarity directly from labeled data~\cite{liu2019roberta,devlin2019bert,he2020deberta}. However, in realistic settings where titles are noisy and incomplete, these methods still struggle on hard cases that require reasoning over product attributes and commercial context rather than surface-level similarity~\cite{seo2025question}.

Large language models (LLMs) offer a complementary approach~\cite{zhao2023survey,chang2024survey,touvron2023llama,achiam2023gpt}. Zero-shot and few-shot prompting can capture semantic correspondence beyond token overlap~\cite{wei2022chain}, while retrieval-augmented methods can ground decisions in external evidence~\cite{lewis2020retrieval,gao2023retrieval}. Multi-agent pipelines further improve robustness by decomposing the decision into specialized sub-tasks (e.g., evidence gathering, attribute comparison, and decision aggregation)~\cite{li2024survey,seo2025question,talebirad2023multi}. Despite their strong performance, these systems introduce practical drawbacks: they often depend on repeated calls to expensive external APIs, require retrieval and multi-step orchestration at inference time (increasing latency and complexity)~\cite{cemri2025multi,li2024survey}, and are difficult to deploy in privacy-sensitive environments that require on-premise operation and predictable cost~\cite{jiang2020can,duan2023flocks}.

To address these issues, we propose \textbf{EPM-RL}, a reinforcement-learning-based framework for accurate and efficient on-premise product mapping. Our key idea is to distill high-cost agentic reasoning into a trainable in-house model. Starting from a human-labeled dataset of product pairs, we first apply parameter-efficient fine-tuning (PEFT)~\cite{hu2022lora,dettmers2023qlora,liu2024dora} with structured reasoning traces so the model learns interpretable comparison behavior. We then refine the model with reinforcement learning~\cite{rafailov2023direct,liu2024deepseek} using verifiable rewards and judge-based signals produced by three specialized agents, and further incorporate unlabeled data to improve reasoning quality under an LLM-as-a-judge objective. In this way, \textbf{EPM-RL} internalizes structured reasoning while avoiding inference-time agent orchestration.

Experiments on an internal product mapping benchmark show that \textbf{EPM-RL} consistently improves over PEFT-only training and offers a better quality--cost trade-off than commercial API-based baselines. In particular, it achieves stronger performance on difficult product pairs while remaining suitable for scalable private deployment. These results suggest that reinforcement learning can turn product mapping from a high-latency agentic pipeline into a production-ready, inspectable, and cost-efficient in-house system.

The main contributions of this paper are as follows:
\begin{itemize}[leftmargin=1.5em]
    \item We propose \textbf{EPM-RL}, an on-premise reinforcement-learning framework for scalable e-commerce product mapping.
    \item We introduce a practical training pipeline that combines PEFT, RL with verifiable and agentic rewards, and additional unlabeled-data optimization with structure and judge-based supervision.
    \item We show that \textbf{EPM-RL} improves over PEFT-only training and provides a stronger quality--cost trade-off than commercial API-based baselines on internal benchmarks.
\end{itemize}

\section{Background}

\paragraph{Task Description.}
Product mapping in e-commerce aims to determine whether a \emph{base product} listing and a \emph{compared product} listing refer to the same underlying real-world product. Formally, given a base product $p_b$ and a compared product $p_c$, the goal is to predict a binary label $y \in \{0,1\}$, where $y=1$ indicates that the two listings correspond to the same product and $y=0$ otherwise. Although the formulation is simple, the task is challenging in real marketplaces because product titles are often noisy, incomplete, and optimized for search visibility rather than standardized description. Sellers may inject promotional keywords, platform-specific tags, origin descriptors, or bundle information, causing the same product to appear under different names or, conversely, making different products appear deceptively similar. Table~\ref{tab:background_examples} illustrates these cases: in the positive examples, the compared listing adds non-essential descriptors (e.g., origin or packaging) while preserving the core product identity; in the hard negative examples, the titles overlap heavily but differ in bundle composition or variant attributes, leading to a different underlying product. These examples illustrate why effective product mapping requires reasoning over attributes such as brand, variant, specification, quantity, and bundle structure rather than relying solely on surface-level lexical overlap.

\begin{table}[t]
\centering
\caption{Illustrative examples of product pairs in e-commerce product mapping.}
\label{tab:background_examples}
\footnotesize
\setlength{\tabcolsep}{3pt}
\renewcommand{\arraystretch}{1.05}
\begin{tabularx}{\columnwidth}{>{\raggedright\arraybackslash}p{1.55cm} >{\raggedright\arraybackslash}X >{\raggedright\arraybackslash}X}
\toprule
\textbf{Case} & \textbf{Base Product} & \textbf{Compared Product} \\
\midrule
\rowcolor{black!10}\multicolumn{3}{l}{\textit{Positive} ($y{=}1$)} \\
& MegaDoseD Vitamin D3 4000IU, 120 tablets $\times$ 3 
& MegaDoseD Vitamin D3 4000IU Swiss-made, 120 tablets $\times$ 3 \\
& Apple AirPods Pro (2nd Gen) with MagSafe Case
& Apple AirPods Pro 2 (2nd generation) MagSafe Charging Case \\
\midrule
\rowcolor{black!10}\multicolumn{3}{l}{\textit{Hard Negative} ($y{=}0$)} \\
& Coca-Cola Zero 355mL $\times$ 24 cans 
& Coca-Cola Zero 355mL $\times$ 24 cans + cooler bag gift set \\
& Nespresso Vertuo Espresso Pods Variety Pack, 30 count
& Nespresso Vertuo Espresso Pods Variety Pack, 60 count \\
\bottomrule
\end{tabularx}
\end{table}

\paragraph{LLM-based Product Mapping.}
Large language models (LLMs) are well suited for product mapping because they produce outputs autoregressively, assigning a probability to each next token conditioned on the input. Let $x=[p_b; p_c]$ denote the concatenated input pair, and let $o=(o_1,\dots,o_{|o|})$ denote the generated output sequence. An autoregressive LLM with parameters $\theta$ defines
\begin{equation}
P_{\theta}(o \mid x)
=
\prod_{t=1}^{|o|} P_{\theta}(o_t \mid x, o_{<t}).
\end{equation}

In practice, there are two common ways to use this token-level distribution for binary product mapping.

\noindent\textbf{(1) Direct classification by label logits.} The model is prompted to output a single label token (e.g., \texttt{0} for non-match and \texttt{1} for match). Let $z_{0}(x)$ and $z_{1}(x)$ denote the pre-softmax logits for the first generated label token. The prediction is obtained by
\begin{equation}
\hat{y} = \arg\max_{y \in \{0,1\}} z_{y}(x)
\quad\Leftrightarrow\quad
\hat{y} = \arg\max_{y \in \{0,1\}} P_{\theta}(y \mid x).
\end{equation}
This approach is simple and efficient but provides no structured explanation.

\noindent\textbf{(2) Reasoning-then-label generation.} Alternatively, the model generates an explicit reasoning trace $r=(r_1,\dots,r_T)$ followed by a final label $y$. The joint probability factorizes as
\begin{equation}
P_{\theta}(r,y \mid x)
=
\left(\prod_{t=1}^{T} P_{\theta}(r_t \mid x, r_{<t})\right)
P_{\theta}(y \mid x, r),
\end{equation}
and the predicted label is extracted from the generated output, e.g.,
\begin{equation}
\hat{y} = \mathrm{parse}\!\left(\arg\max_{o} P_{\theta}(o \mid x)\right),
\end{equation}
where $\mathrm{parse}(\cdot)$ returns the value inside the \texttt{<label>\dots</label>} field.

In \textbf{EPM-RL}, we adopt the reasoning-then-label formulation with a fixed structured format ({<reason>\dots</reason><label>\dots</label>). This choice improves interpretability by exposing an explicit comparison rationale, enables supervised learning of structured reasoning traces during PEFT, and supports fine-grained reward shaping in GRPO that separately evaluates format compliance, label correctness, and judge-based reasoning quality.

\section{Methodology}

\subsection{Data Preprocessing.}
Our raw dataset consists of triplets $(p_b, p_c, y)$, where $p_b$ denotes the base product title, $p_c$ denotes the compared product title, and $y \in \{0,1\}$ is the human-provided binary label indicating whether the two listings refer to the same underlying product. We first partition the dataset into four disjoint subsets for PEFT training, reinforcement learning, validation, and testing. To further support reasoning-aware training, we construct an additional reflective reasoning dataset from the PEFT split. Specifically, given $(p_b, p_c, y)$, we prompt a strong LLM to generate a reasoning trace that explains why the pair should be judged as matched or unmatched, while assuming a \emph{blinded} setting in which the model must reason only from the information available in the two product titles. This reverse generation process produces structured reasoning traces aligned with the human label and yields augmented tuples $(p_b, p_c, r, y)$, where $r$ denotes the synthesized reflective reasoning.

\subsection{Parameter-Efficient Fine-Tuning.}
After constructing the reasoning-augmented data, we format each training instance as an input-output pair. The input contains the base prompt together with the base product and compared product, while the target output contains both the reasoning trace and the final binary label in a structured format, namely
<reason>...</reason><label>...</label>.
This format encourages the model to learn not only the final decision but also the intermediate comparison logic. We then optimize the model with parameter-efficient fine-tuning (PEFT)~\cite{hu2022lora}, updating only a small subset of trainable parameters while keeping the backbone model largely frozen. Let $x=[p_b; p_c]$ denote the input and let $o=(r,y)$ denote the target output sequence. The PEFT objective follows the standard autoregressive language modeling loss:
\begin{equation}
\mathcal{L}_{\mathrm{PEFT}}(\theta)
=
-\sum_{i=1}^{N}\sum_{t=1}^{T_i}
\log P_{\theta}\!\left(o_{i,t}\mid x_i, o_{i,<t}\right),
\end{equation}
where $\theta$ denotes the trainable PEFT parameters, $N$ is the number of training instances, and $T_i$ is the output length of the $i$-th example. Through this stage, the model learns to generate reasoning traces and labels jointly from product-pair inputs.

\subsection{Reinforcement Learning with GRPO.}
Although PEFT teaches the model a structured reasoning format, it does not explicitly optimize the quality of the generated reasoning or its alignment with the final decision. To address this, we further refine the model with Group Relative Policy Optimization (GRPO)~\cite{liu2024deepseek}. Given an input $x$, the current policy $\pi_{\theta}$ generates a group of $K$ rollout responses $\{o^{(1)},\dots,o^{(K)}\}$. Each rollout is assigned a scalar reward based on both verifiable signals and LLM-based judge signals, and the policy is updated to increase the likelihood of better-performing rollouts relative to others in the same group. Let $r^{(k)}$ denote the reward for the $k$-th rollout and let $\hat{A}^{(k)}$ denote its normalized relative advantage within the group. The GRPO objective can be written as
\begin{equation}
\mathcal{L}_{\mathrm{GRPO}}(\theta)
=
-\mathbb{E}_{x}\left[
\frac{1}{K}\sum_{k=1}^{K}
\hat{A}^{(k)}
\log \pi_{\theta}\!\left(o^{(k)} \mid x\right)
\right].
\end{equation}

In our setting, the reward function combines a verifiable reward (VR) with agentic reward scores. The verifiable part checks whether the output follows the required structure and whether the predicted label is valid. The agentic part is computed by three specialized LLM judges that evaluate complementary aspects of the reasoning: (1) a \emph{Core Identity} judge that assesses whether the reasoning correctly identifies and compares the central product/category identity, (2) a \emph{Model-Identifier} judge that evaluates whether the reasoning properly handles brand, model line, and explicit identifier tokens, and (3) a \emph{Variant-Conflict} judge that measures how well the reasoning detects variant-level consistency or conflict, such as size, capacity, count, option, or bundle differences. Denoting these scores by $s_{\mathrm{ver}}$, $s_{\mathrm{core}}$, $s_{\mathrm{id}}$, and $s_{\mathrm{var}}$, respectively, the overall reward is defined as
\begin{equation}
R(o,x)
=
\lambda_{\mathrm{ver}} s_{\mathrm{ver}}
+
\lambda_{\mathrm{core}} s_{\mathrm{core}}
+
\lambda_{\mathrm{id}} s_{\mathrm{id}}
+
\lambda_{\mathrm{var}} s_{\mathrm{var}},
\end{equation}
where $\lambda_{\mathrm{ver}}$, $\lambda_{\mathrm{core}}$, $\lambda_{\mathrm{id}}$, and $\lambda_{\mathrm{var}}$ are weighting coefficients. This design allows the model to improve not only final classification accuracy but also the quality and faithfulness of its reasoning process.

\section{Experiments}
\subsection{Setup}
\subsubsection{Dataset Used}
We evaluate our framework on an internally collected and curated product mapping dataset consisting of approximately 12K labeled product pairs. Each instance contains a \emph{base product}, a \emph{compared product}, and a binary human label indicating whether the two listings refer to the same underlying product. The dataset covers approximately 500 brands collected from diverse e-commerce platforms, reflecting substantial variation in naming conventions, promotional expressions, bundle descriptions, and platform-specific title formats. This diversity makes the benchmark representative of realistic product-mapping scenarios encountered in practice.

To support different stages of training and evaluation, we partition the dataset into four disjoint splits:
\begin{itemize}[leftmargin=1.5em]
    \item 6K examples for parameter-efficient fine-tuning (PEFT),
    \item 4K examples for reinforcement learning (RL),
    \item 1K examples for validation,
    \item 1K examples for testing.
\end{itemize}
The splits are constructed using stratified sampling to preserve both brand diversity and label balance across subsets, thereby reducing sampling bias and ensuring a fair comparison between training and evaluation stages. In particular, we maintain a balanced distribution of positive and negative labels across all splits. In the full dataset, positive cases account for \textbf{70.6\%} of the instances, while negative cases account for \textbf{29.4\%}. This stratified design helps ensure that the model is exposed to a broad range of matching and non-matching patterns during training while being evaluated on a similarly representative test distribution.

\subsubsection{Baselines}
To evaluate the effectiveness of our proposed framework, we compare \textbf{EPM-RL} against representative baselines spanning encoder-based binary classification, single-inference prompting, retrieval-augmented inference, and agentic multi-step pipelines:
\begin{itemize}[leftmargin=1.5em]
    \item \textbf{Encoder binary classifier}~\cite{devlin2019bert,liu2019roberta}: A supervised baseline that trains a text encoder to directly classify whether $(p_b,p_c)$ match. We encode the concatenated pair and train a lightweight classification head with cross-entropy loss.

    \item \textbf{Zero-shot}: A direct single-inference baseline where the LLM receives the base product and compared product titles and predicts the binary match label without any intermediate structure.

    \item \textbf{Entity Attribution}: A single-inference baseline that encourages the LLM to explicitly extract and compare salient attributes (e.g., brand, product type, specification, quantity, and option-related cues) before predicting the label.
    
    \item \textbf{Chain-of-Thought (CoT)}~\cite{wei2022chain}: A single-inference baseline that prompts the LLM to generate a step-by-step rationale before producing the final binary decision.

    \item \textbf{RAG}~\cite{lewis2020retrieval}: A retrieval-augmented generation baseline that retrieves evidence using BM25 and appends it to the prompt before prediction. The BM25 query is constructed by concatenating the two titles as \texttt{base product [SEP] compared product}.

    \item \textbf{Multi-Agent RAG}~\cite{althaf2025multi}: An agentic retrieval baseline that decomposes inference into three steps: a \emph{direct} agent predicts a label from the titles only, an \emph{indirect} agent predicts a label conditioned on BM25-retrieved evidence, and a \emph{coordinator} agent aggregates the two predictions into a final decision.
\end{itemize}

\noindent\textbf{Encoder classifier formulation.} Let $x=[\mathrm{[CLS]};p_b;\mathrm{[SEP]};p_c;\mathrm{[SEP]}]$ be the packed title pair and let $h=\mathrm{Enc}(x)_{\mathrm{[CLS]}}\in\mathbb{R}^d$ be the encoder's pooled representation. We predict the match probability with a logistic head:
\begin{equation}
\hat{p}(y{=}1\mid x)=\sigma\left(w^{\top}h+b\right),
\end{equation}
where $w\in\mathbb{R}^d$, $b\in\mathbb{R}$ are trainable parameters and $\sigma(\cdot)$ is the sigmoid function. The model is trained by minimizing the binary cross-entropy:
\begin{equation}
\mathcal{L}_{\mathrm{enc}}=-\,y\log \hat{p}-(1-y)\log(1-\hat{p}).
\end{equation}

The first three baselines belong to the \emph{single-inference} setting. In contrast, \textit{RAG} and \textit{Multi-Agent RAG} incorporate external evidence and multi-step reasoning, providing stronger but higher-latency inference-time baselines.

For fairness, all baselines share the same base instruction prompt, and differ only in whether they add structured reasoning, retrieved evidence, or agentic orchestration.

\subsubsection{Models Used}

\paragraph{Encoder Models (Binary Classification)}
We train and compare four encoder-based binary classifiers: \textbf{BERT-base}, \textbf{BERT-large}~\cite{devlin2019bert}, \textbf{RoBERTa-base}, and \textbf{RoBERTa-large}~\cite{liu2019roberta}. Each model is implemented as a cross-encoder over the packed title pair and trained with the binary cross-entropy objective described in the baselines section. We train for up to $5$ epochs with early stopping (patience $2$), using a batch size of $1024$ and a learning rate of $1\times 10^{-5}$.

\paragraph{Large Language Models}
We use models from the (1) \textbf{Nemotron-Nano-3} family~\cite{blakeman2025nemotron} and (2) \textbf{GPT-5.4 reasoning}, assigning each to a distinct role in the pipeline. As a strong teacher and reference baseline, we use \textbf{Nemotron-Nano-3 Super-120B}\footnote{\url{https://huggingface.co/nvidia/NVIDIA-Nemotron-3-Super-120B-A12B-BF16}} for (i) synthesizing reflective reasoning traces for PEFT training and (ii) reporting a high-capacity inference-time baseline. For reinforcement learning, we use \textbf{GPT-5.4 reasoning}\footnote{\url{https://openai.com/index/introducing-gpt-5-4}} to implement the three judge agents that score reasoning quality. Our main trainable and evaluation model is \textbf{Nemotron-Nano-3 30B-A3B}\footnote{\url{https://huggingface.co/nvidia/NVIDIA-Nemotron-3-Nano-30B-A3B-BF16}} (referred to as \textbf{Nemotron-Nano-3-30B}), a hybrid Mixture-of-Experts model with 30B total parameters and roughly 3.5B active parameters per token.

Across all experiments, decoding uses temperature $0.7$ and top-$p$ $0.95$, with a maximum output length of $1024$ tokens to accommodate both titles and structured reasoning outputs.

\paragraph{Retrieval Model}
For retrieval-based baselines, we use \textbf{BM25}~\cite{robertson2009probabilistic} to retrieve the top-$k{=}5$ evidence candidates from an internal retrieval corpus. Given a query $q$ (constructed from the base product and compared product titles) and a candidate document $d$, BM25 computes a relevance score
\begin{equation}
\mathrm{BM25}(q,d)
=
\sum_{t \in q}
\mathrm{IDF}(t)
\cdot
\frac{f(t,d)\,(k_1+1)}{f(t,d)+k_1\left(1-b+b\,\frac{|d|}{\mathrm{avgdl}}\right)},
\end{equation}
where $\mathrm{IDF}(t)$ denotes the inverse document frequency of term t, measuring how rare or informative t is across the document collection, $f(t,d)$ is the term frequency of token $t$ in $d$, $|d|$ is the document length, and $\mathrm{avgdl}$ is the average document length in the corpus. We use the default parameters $k_1{=}1.2$ and $b{=}0.75$. The retrieved evidence is appended to the LLM prompt for the \textit{RAG} and \textit{Multi-Agent RAG} baselines described above.

\subsubsection{PEFT and RL Setups}

\paragraph{PEFT Setup.}
For parameter-efficient fine-tuning (PEFT), we use \textbf{Low-Rank Adaptation (LoRA)}~\cite{hu2022lora} to adapt the model by learning low-rank updates to selected linear projections while keeping the backbone weights frozen. Given a pretrained projection matrix $W \in \mathbb{R}^{d \times k}$, LoRA parameterizes the update as
\begin{equation}
W' = W + \Delta W, \qquad \Delta W = BA,
\end{equation}
where $A \in \mathbb{R}^{r \times k}$ and $B \in \mathbb{R}^{d \times r}$ are trainable low-rank matrices and $r \ll \min(d,k)$ is the LoRA rank. With the standard scaling rule, the adapted projection becomes
\begin{equation}
W' = W + \frac{\alpha}{r}BA,
\end{equation}
where $\alpha$ is a scaling factor.

To make the adaptation lightweight, we apply LoRA \emph{only} to the attention projections. For each transformer layer $\ell$, we update
\begin{equation}
W_{q}^{(\ell)}\!,\; W_{k}^{(\ell)}\!,\; W_{v}^{(\ell)}
\quad\text{with LoRA, and keep all other parameters frozen.}
\end{equation}

We set the learning rate to $1\times10^{-5}$, batch size to $4$, LoRA rank to $r{=}32$, and scaling factor to $\alpha{=}64$, and train for $5$ epochs. This configuration provides a practical trade-off between parameter efficiency and adaptation capacity while keeping training stable and computationally manageable.

\paragraph{RL Setup.}
\textbf{RL parameters.} After PEFT, we further optimize the model with reinforcement learning using \textbf{Group Relative Policy Optimization (GRPO)}~\cite{liu2024deepseek}. In this stage, we keep the same adapter configuration as in PEFT so that the RL update remains parameter-efficient and directly refines the previously learned reasoning behavior. As in the PEFT stage, we freeze the backbone model and train only the LoRA adapter parameters. Specifically, the LoRA rank and scaling factor are kept unchanged, and updates are again restricted to the adapter parameters. We set the learning rate for RL to $5\times10^{-5}$, the training batch size to $4$, and the number of rollouts per input to $4$. To stabilize optimization, we use a maximum clipping value of $0.1$ and apply a dropout rate of $0.05$. The model is trained for $1$ epoch.

\textbf{Reward design.} We define the total reward as a weighted combination of three components: (i) \emph{structured output compliance} (weight $1$), which checks whether the model output follows the required XML-like format (e.g., \texttt{<reason>} and \texttt{<label>} fields); (ii) \emph{binary decision correctness} (weight $2$), which assigns reward based on whether the predicted binary label matches the ground-truth label; and (iii) \emph{LLM-as-a-judge preference} (weight $1$), computed by averaging the scores from three specialized judge agents.

Formally, let $s_{\mathrm{fmt}}\in\{0,1\}$ denote the format-compliance score, $s_{\mathrm{cls}}\in\{0,1\}$ denote the label-correctness score, and let $s_{\mathrm{judge}} = \frac{1}{3}\sum_{j=1}^{3} s_j$ with $s_j\in[0,1]$ be the mean judge score. Let weights $\lambda_{\mathrm{fmt}}{=}1$, $\lambda_{\mathrm{cls}}{=}2$, and $\lambda_{\mathrm{judge}}{=}1$. The normalized reward is
\begin{equation}
R
=
\frac{\sum\limits_{u\in\{\mathrm{fmt},\mathrm{cls},\mathrm{judge}\}} \lambda_u\, s_u}{\sum\limits_{u\in\{\mathrm{fmt},\mathrm{cls},\mathrm{judge}\}} \lambda_u}
\in [0,1].
\end{equation}

Compared with PEFT, this stage focuses less on imitation of target outputs and more on refining generation behavior according to reward signals, enabling the model to improve the quality, consistency, and decision usefulness of its reasoning traces under the GRPO objective.

\subsubsection{Hardware / Framework Specification}
We conducted LoRA fine-tuning of NVIDIA’s Nemotron model using the Megatron-Bridge\footnote{\url{https://github.com/NVIDIA-NeMo/Megatron-Bridge}} SFT framework on a node equipped with 8 NVIDIA H100 GPUs. Subsequently, we performed LoRA-based reinforcement learning with Hugging Face and TRL using the same GPU configuration. This setup allowed us to efficiently adapt the model during training stages.

\subsection{Judge Agent Design}
\label{sec:judge-agent-design}

We implement the LLM-as-a-judge component in GRPO with three specialized judge agents. These judges are motivated by a small expert analysis: a human domain expert inspected 100 product-pair samples and identified three recurring reasoning patterns (core identity matching, model/identifier matching, and variant-level conflict checking) that most strongly determined correct decisions. Accordingly, each judge takes as input the \emph{base product} title $p_b$, the \emph{compared product} title $p_c$, and the model-generated reasoning text $r$ extracted from \texttt{<reason>\dots</reason>}, and outputs a single scalar score $s\in[0,1]$ (one float with no explanation) for its designated sub-skill; the scores are later aggregated into the judge-based reward term. To reduce reward hacking and improve faithfulness, we instruct judges to (a) evaluate only their designated sub-skill, (b) penalize hallucinated tokens not present in the input titles, and (c) return $0.0$ when the extracted reasoning is missing or empty.

\paragraph{Core Identity agent (Step-1).}
This agent scores whether the reasoning correctly identifies and compares the \emph{core product/category identity} shared or mismatched between the base and compared titles. It focuses on the central product type and key anchor tokens (e.g., the main product/category terms), and ignores brand/model codes and variant attributes. Scores are assigned on a graded rubric from generic or weak core comparison ($\approx 0.5$) to token-grounded and correct core identification ($1.0$).

\paragraph{Model-Identifier agent (Step-2).}
This agent scores whether the reasoning correctly handles \emph{brand and model identifiers}, including brand prefixes, model lines, and explicit model codes. It evaluates whether the reasoning distinguishes ignorable prefixes from true identifiers and grounds identifier comparisons in tokens appearing in the base and compared titles. It does not judge variant conflicts (e.g., size/count differences).

\paragraph{Variant-Conflict agent (Step-3).}
This agent scores whether the reasoning checks \emph{variant attributes} and detects conflicts or consistency between the base and compared titles. It focuses on attributes such as size, color, capacity, count, specification, option, bundle composition, and version. The score reflects how completely and correctly the reasoning cites relevant tokens and determines whether variant-level differences imply a mismatch.

\begin{table}[t]
\centering
\caption{Prompt templates used in our experiments. We use a shared Baseline prompt for single-inference baselines and a structured PEFT prompt for supervised fine-tuning.}
\label{tab:prompts}
\footnotesize
\setlength{\tabcolsep}{6pt}
\renewcommand{\arraystretch}{1.05}

\begin{tabularx}{\columnwidth}{X}
\toprule
\textbf{Baseline}\\
\midrule
\begin{minipage}[t]{\linewidth}
You are a product matching classifier. Decide whether Product A and Product B refer to the same sellable product. Output only 1 if matched, otherwise output only 0.
\begin{enumerate}[leftmargin=1.2em, nosep]
  \item Different product family, series, or model name means 0.
  \item Different variant attributes that change the sellable SKU (such as color, size, capacity, count, edition, generation, set composition, pack size) means 0.
  \item Ignore only minor formatting differences, spacing, punctuation, obvious spelling noise, or packaging-only phrases.
  \item If uncertain, output 0.
\end{enumerate}
Output only one character: 0 or 1.
\end{minipage}
\\
\midrule
\textbf{PEFT}\\
\midrule
\begin{minipage}[t]{\linewidth}
You are a product-title matching analyst.

\textbf{Task:} Given two product names, decide whether they refer to the same sellable product variant.
\begin{itemize}[leftmargin=1.2em, nosep]
  \item Product A: \{p1\_name\}
  \item Product B: \{p2\_name\}
\end{itemize}

\textbf{Important constraints:}
\begin{enumerate}[leftmargin=1.2em, nosep]
  \item Do NOT use any external label column or hidden metadata. Decide only from the two product names.
  \item Compare step by step before concluding.
  \item Treat differences in model code, capacity, size, color, quantity, option, bundle composition (set/single item), edition/origin/version as potentially critical.
  \item Ignore minor wording differences such as spacing, punctuation, seller prefix/brand prefix, and marketing words when core identity is still the same.
  \item If core product identity or key variant conflicts, output 0.
  \item If core product identity and key variant are consistent (or one side is only less specific without contradiction), output 1.
\end{enumerate}

\textbf{Output format (must follow exactly):}
\begin{enumerate}[leftmargin=1.2em, nosep]
  \item First, <identify the core product/category and main tokens in both names>.
  \item Second, <compare brand/model line/model number and key identifiers>.
  \item Third, <compare variant attributes: size/color/count/spec/option/bundle, and check for conflicts>.
  \item So, the final answer is: <0 or 1>.
\end{enumerate}

\textbf{Label meaning:}
\begin{itemize}[leftmargin=1.2em, nosep]
  \item 1 = matched (same sellable product variant)
  \item 0 = not matched (different product or conflicting variant)
\end{itemize}

\textbf{Expected output:} <reason>evidence</reason><label>0/1</label>
\end{minipage}
\\
\bottomrule
\end{tabularx}
\end{table}

\begin{table*}[t]
  \centering
  \normalsize
  \caption{Main experiment results on the internal product-mapping test set. We report Accuracy (Acc), Precision (Prec), Recall (Rec), and F1. \emph{Single inference} uses one-pass prompting without external tools; \emph{RAG} augments the prompt with BM25-retrieved evidence; \emph{Multi-Agent RAG} coordinates multiple retrieval-and-reasoning agents over the same BM25 evidence pool. \emph{Ours} summarizes our parameter-efficient variants (LoRA + Reasoning; LoRA + GRPO Reasoning). For each metric, \textbf{bold} indicates the best score and \underline{underline} indicates the second best score.}
  \label{tab:main-exp-results}
  \setlength{\tabcolsep}{9pt}
  \renewcommand{\arraystretch}{1.25}
  \begin{tabular}{lcccc}
    \toprule
    Method & Accuracy $\uparrow$ & Precision $\uparrow$ & Recall $\uparrow$ & F1-Score $\uparrow$ \\
    \midrule

    \rowcolor{black!10}\multicolumn{5}{l}{\textit{Encoder Binary Classifier}} \\
    BERT-base  & 0.8250 & 0.6814 & 0.8235 & 0.7458 \\
    BERT-large & 0.8290 & 0.6713 & 0.8197 & 0.7381 \\
    RoBERTa-base & 0.8367 & 0.7405 & 0.7326 & 0.7366 \\
    RoBERTa-Large & 0.8300 & 0.6741 & 0.8163 & 0.7385 \\
    
    \rowcolor{black!10}\multicolumn{5}{l}{\textit{Single inference}} \\
    Zero Shot Inference          & 0.8540 & 0.8109 & 0.6565 & 0.7256 \\
    Chain-of-Thought Reasoning   & 0.8450 & 0.8639 & 0.5612 & 0.6804 \\
    Entity-Attribute Reasoning   & 0.8440 & 0.8750 & 0.5476 & 0.6736 \\

    \rowcolor{black!10}\multicolumn{5}{l}{\textit{RAG}} \\
    Retrieval-Augmented Generation (BM25) & 0.8490 & 0.7227 & 0.7891 & 0.7545 \\

    \rowcolor{black!10}\multicolumn{5}{l}{\textit{Multi-Agent RAG}} \\
    Multi-Agent Retrieval-Augmented Generation (BM25) & \textbf{0.8640} & \underline{0.8911} & 0.6122 & 0.7258 \\

    \rowcolor{black!10}\multicolumn{5}{l}{\textit{Reasoning LLM}} \\
    GPT-5.4 reasoning                          & \underline{0.8630} & \textbf{0.9758} & 0.5476 & 0.7015 \\

    \rowcolor{black!10}\multicolumn{5}{l}{\textit{Ours}} \\
    LoRA + Reasoning        & 0.8570 & 0.7103 & \underline{0.8673} & \underline{0.7810} \\
    LoRA + GRPO Reasoning   & 0.8450 & 0.7815 & \textbf{0.8735} & \textbf{0.8120} \\

    \bottomrule
  \end{tabular}
\end{table*}

\subsection{Prompts}
\label{sec:prompts}
We provide the main prompt templates used in our experiments: a shared \textbf{Baseline} prompt for all single-inference baselines and a structured \textbf{PEFT} prompt for supervised fine-tuning in Table~\ref{tab:prompts}.

The \textbf{Baseline} prompt and is kept identical across baselines so that any performance differences are attributable to the inference strategy rather than prompt phrasing. In contrast, the \textbf{PEFT} prompt elicits a structured three-stage comparison---\emph{First} (core product/category identity), \emph{Second} (brand/model identifiers), and \emph{Third} (variant attributes and bundle composition)---before asking for a final binary decision.

This ``First/Second/Third'' decomposition mirrors the competencies evaluated by our three judge agents (Core Identity, Model-Identifier, and Variant-Conflict; Section~\ref{sec:judge-agent-design}). As a result, the fine-tuned model learns to produce reasoning traces that are directly scorable by the judges, which improves the stability and interpretability of reward shaping during GRPO. Finally, using a shared Baseline prompt and a judge-aligned PEFT prompt reduces prompt-induced variance across training and evaluation, helping ensure that improvements reflect model capability rather than formatting artifacts.

\subsection{Main Experiement Results}
We compare \textbf{EPM-RL} against a diverse set of baselines, including encoder-based binary classifiers, single-inference prompting methods, retrieval-augmented approaches, multi-agent reasoning pipelines, and strong large language models. As shown in Table 3, the single-inference prompting baselines provide a useful starting point but exhibit clear limitations. Zero-shot inference achieves the strongest overall balance among the prompting-based variants, while Chain-of-Thought and Entity-Attribute Reasoning do not consistently improve performance despite introducing more explicit intermediate reasoning. We additionally include encoder binary classifiers, such as BERT and RoBERTa, as representative non-generative baselines to contextualize the gains from reasoning-centric approaches. Overall, these results suggest that prompting the model to reason more verbosely does not necessarily translate into better product mapping performance, especially when the decision depends on subtle distinctions in bundle composition, variant conflict, or product identity.

Among the stronger inference-time baselines, retrieval-based and agentic methods show a distinct trade-off. Standard RAG improves recall and achieves the strongest F1 score among the non-agentic baselines, indicating that retrieved evidence is helpful for resolving ambiguous product pairs. Multi-Agent RAG further strengthens direct comparison through coordinated evidence-based reasoning and achieves the highest accuracy and precision among the listed baselines. However, its recall remains relatively low, suggesting that the multi-agent pipeline tends to make more conservative positive predictions. One possible explanation is that the multi-agent setting amplifies conservative decisions through agent coordination and voting: when multiple agents independently assess the same pair, disagreement or uncertainty may be resolved toward non-match decisions, thereby increasing precision at the cost of recall. In addition, because the retrieval corpus and evidence construction are derived from the same internal product-mapping environment, Multi-Agent RAG may partially reflect label-distribution or decision-pattern biases present in the training data, which can further contribute to its high precision.

A similar tendency is observed for the GPT-5.4 reasoning baseline, which achieves extremely high precision but substantially lower recall. This pattern suggests that the model is highly selective in predicting positive matches. Rather than indicating uniformly superior matching ability, the result may reflect an inherent decision bias in the reasoning model toward avoiding false positives under ambiguous product-title comparisons. In product mapping, such conservatism can be beneficial when precision is prioritized, but it can also lead to many missed matches, as reflected in the lower recall and F1 score. These observations highlight that high precision alone is insufficient to characterize overall product-mapping quality; a practical system must also maintain enough recall to recover valid matches across noisy and heterogeneous marketplace titles.

In contrast, our framework \textbf{EPM-RL} is designed to internalize structured reasoning into the model itself rather than depending on repeated inference-time coordination. The PEFT-based variant already provides a stronger and more practical alternative to prompt-only baselines by teaching the model to generate product-aware reasoning traces together with the final decision. Building on this, the GRPO-enhanced variant further improves the faithfulness and usefulness of the generated reasoning through verifiable rewards and specialized judge-based reward signals. As a result, \textbf{EPM-RL} achieves the best overall F1 performance on the product-mapping benchmark, outperforming both single-inference and retrieval-based baselines in terms of balanced decision quality while remaining more suitable for efficient on-premise deployment. In particular, its higher recall indicates that the model can recover substantially more true matches than highly conservative baselines, while its precision remains competitive. These findings suggest that distilling agentic reasoning into a trainable in-house model is more effective than relying solely on prompt engineering, large proprietary reasoning models, or costly inference-time retrieval pipelines.

Overall, the results demonstrate three key observations. First, purely prompt-based reasoning is not sufficient for robust product mapping in realistic e-commerce settings. Second, retrieval- and agent-based methods can improve grounding and precision, but their benefits may come with conservative decision behavior, label-distribution sensitivity, and additional system overhead. Third, \textbf{EPM-RL} offers a more favorable balance by combining reasoning quality, predictive performance, and deployment efficiency within a unified trainable framework.

\section{Conclusion}
We introduced \textbf{EPM-RL}, a reinforcement-learning framework for building an accurate and efficient on-premise product mapping model. Our approach distills high-cost agentic reasoning into a trainable in-house model by combining parameter-efficient fine-tuning with structured reasoning traces and GRPO-based reinforcement learning using verifiable rewards and judge-agent signals. Our results on an internal product-mapping benchmark demonstrate that \textbf{EPM-RL} can improve over prompt-only and retrieval-heavy baselines while avoiding inference-time orchestration.

\section{Limitations}
\begin{enumerate}[leftmargin=1.5em]
  \item \textbf{Internal benchmark scope.} We have evaluated primarily on an internally curated dataset, and additional experiments on public, cross-domain product-matching benchmarks are needed to better assess generalization.
  \item \textbf{Limited qualitative analysis.} Our paper provides limited qualitative analysis of generated reasoning traces and judge-agent scoring behavior. Deeper case studies would help clarify failure modes, reward sensitivity, and reasoning faithfulness.
  \item \textbf{Incomplete system and ablation coverage.} We do not yet provide a full ablation of reward components, judge prompts, and training hyperparameters, nor a detailed study of latency/cost trade-offs across deployment settings. We will add further analysis in the future.
\end{enumerate}

\bibliographystyle{ACM-Reference-Format}
\bibliography{sample-base}

\end{document}